\documentclass{article}

\usepackage{microtype}
\usepackage{graphicx}
\usepackage{subcaption}
\usepackage{booktabs} 

\usepackage{hyperref}

\usepackage[accepted]{icml2026/icml2026}

\usepackage{amsmath}
\usepackage{amssymb}
\usepackage{mathtools}
\usepackage{amsthm}

\usepackage[capitalize,noabbrev]{cleveref}
\icmltitlerunning{Absorber LLM: Harnessing Causal Synchronization to Internalize Contexts into Transformer Parameters}

\begin{document}

\twocolumn[
  \icmltitle{Absorber LLM: Harnessing Causal Synchronization for Test-Time Training}



  \icmlsetsymbol{equal}{*}

  \begin{icmlauthorlist}
    \icmlauthor{Zhixin Zhang}{equal,yyy}
    \icmlauthor{Shabo Zhang}{equal,yyy}
    \icmlauthor{Chengcan Wu}{equal,yyy}
    \icmlauthor{Zeming Wei}{yyy}
    \icmlauthor{Meng Sun}{yyy}
  \end{icmlauthorlist}

  \icmlaffiliation{yyy}{Peking University}


  \icmlcorrespondingauthor{Zhixin Zhang}{2300010815@stu.pku.edu.cn}


  \vskip 0.3in
]



\printAffiliationsAndNotice{\icmlEqualContribution}

\begin{abstract}
Transformers suffer from a high computational cost that grows with sequence length for self-attention, making inference in long streams prohibited by memory consumption. Constant-memory alternatives such as RNNs and SSMs compress history into states with fixed size and thus lose long-tail dependencies, while methods that memorize contexts into parameters, such as Test-Time Training (TTT), are prone to overfitting token-level projection and fail to preserve the causal effect of context in pretrained LLMs. We propose Absorber LLM, which formulates long-context retention as a self-supervised causal synchronization: after absorbing historical contexts into parameters, a contextless model should match the original model with full context on future generations. We optimize this objective by synchronizing internal behaviors of the updated model with the original one, ensuring context absorption and generalization. Experiments on long-context and streaming benchmarks show that Absorber LLM reduces inference memory and improves accuracy over prior parameter-as-memory baselines.
\end{abstract}
\section{Introduction}

Transformers~\cite{vaswani2017attention} underpin modern AI. However, in real-world deployments, transformer models struggle to satisfy further expectations, such as maintaining lifelong memories of user interactions across multiple domains~\cite{park2023generative, wang2024survey, jimenez2023swe, bairi2024codeplan}, and achieving the ultimate goal of continuously absorbing a never-ending stream of data without resetting their state. This bottleneck stems from their inability to effectively process super-long contextual streams. Specifically, the self-attention mechanism of transformers~\cite{vaswani2017attention, dao2022flashattention} incurs a quadratic computational complexity $\mathcal{O}(N^2)$ to generate texts. 
Although variants like sparse~\cite{beltagy2020longformer} and linearized attention~\cite{katharopoulos2020transformers} mitigate this overhead, they merely delay the inevitable exhaustion of cache and computing resources in long-stream scenarios.

To resolve these memory constraints, prior works have proposed linear-time models, including Recurrent Neural Networks~(RNNs)~\cite{peng2023rwkv} and State Space Models (SSMs)~\cite{gu2021combining, gu2024mamba}. 
These models achieve efficiency by compressing history into fixed-size hidden states. However, recent theoretical and empirical analyses reveal that this efficiency comes at the cost of model expressive power, leading to significant information loss compared to full-attention transformers~\cite{merrill2024illusion, hsieh2024ruler}. 

Alternatively, Test-Time Training (TTT)~\cite{sun2024learning, inplacettt} methods utilize model parameters for context storage to overcome state capacity limits. Specifically, given a sequence $X$ to be stored in the parameters of a model, TTT methods train the model to project $X$ into a specifically constructed target $V$. Such methods optimize the model's performance on $X$ by writing it directly into the model parameters.  
Yet, when learning $X$, they ignore the model's performance on \textbf{subsequent} sequences, failing to help the model \textit{absorb} $X$ in a way that contributes to future inferences.
Crucially, such reconstruction methods overlook the causal mechanisms between historical contexts and subsequent inferences, which are essential for effective model deduction~\cite{geiger2024finding}. 

\begin{figure*}[t!]
\centering
\includegraphics[width=0.9\textwidth]{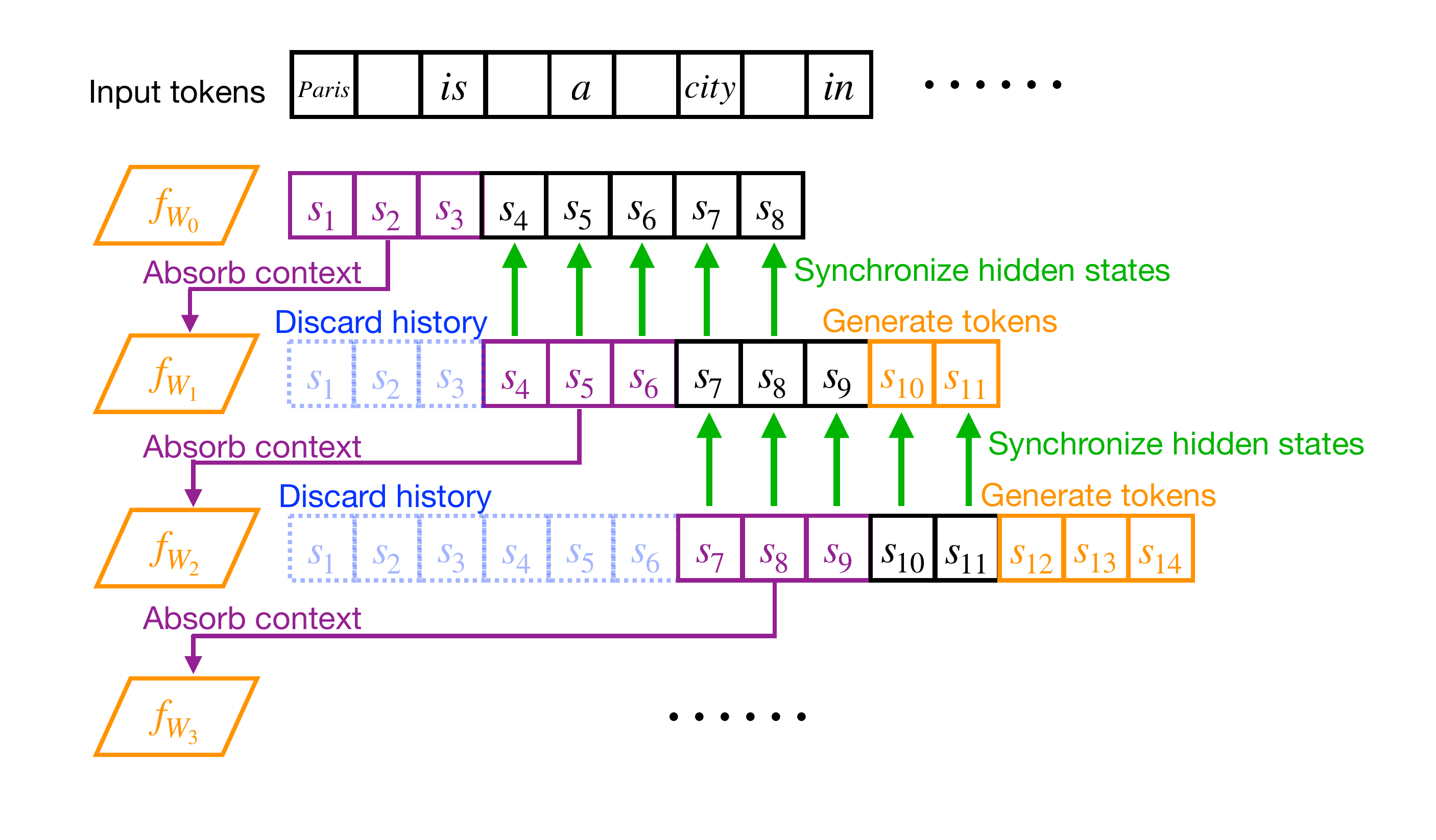} 
\vspace{-20pt}
\caption{A brief overview of our method. We propose to absorb historical contexts and move beyond simple context memorization, focusing on preserving the causal relationship between historical context and future generations rather than directly reconstructing historical tokens. Specifically, we fine-tune the context-less model to behave identically to the original model with full contexts. This is achieved through a self-supervised optimization that synchronizes the hidden states or output behaviors of the two models. This example shows the context absorption process, where the model absorbs 3 historical tokens by synchronization on 5 future tokens. Our method ensures that the absorbed context retains its semantic and causal influence on future deductions.
}
\vspace{-10pt}
\label{fig:overview}
\end{figure*}

In this paper, we draw inspiration from TTT and further propose to preserve the causal relation between historical contexts and the subsequent generation process. We introduce Absorber LLM~(Fig.~\ref{fig:overview}), which ensures causal effect preservation by training the contextless model to behave identically to the full-context model in future inferences. Note that compared to prior TTT methods, our approach focuses on causal effect synchronization rather than simple target projection. We maintain the contribution of removed historical sequences to future inferences, which is crucial for more effective deductions. 

Our main contributions are as follows:
\begin{itemize}
    \item 
    We move beyond directly memorizing contexts to preserving their causal effects on subsequent deductions, and formally define such causal effect preservation as a functional equivalence problem, providing a standard for information transfer from contexts to parameters.
    
    \item 
    We introduce an update mechanism to satisfy the equivalence condition in transformer models, successfully synchronizing the behavior of the updated contextless model with the ideal full-context model, and realizing causal effect preservation and context absorption.
    
    \item 
    We validate the efficiency and effectiveness of our method on long-context benchmarks. Our results demonstrate that Absorber LLM not only outperforms traditional transformer models in computational efficiency but also achieves superior performance compared to prior linear-time and parameter memory methods, paving the way for scalable, infinite-stream inference in real-world deployments.
\end{itemize}

\section{Motivation}

\begin{figure*}
    \centering
    \includegraphics[width=0.7\linewidth]{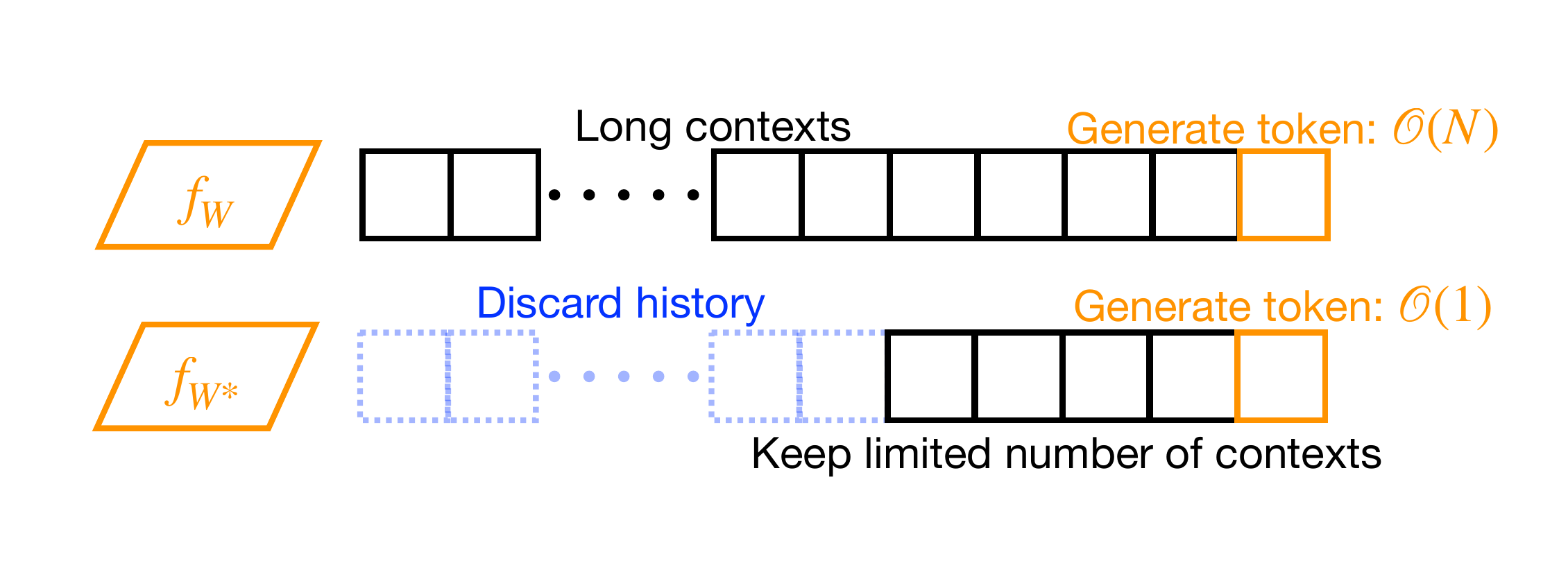}
    \vspace{-10pt}
    \caption{A demonstration of the necessity of inference with a limited number of contexts. Normally, transformers compute $f_W(x_0 x_1 \dots x_{n})$ to generate the next token $x_{n+1}$ with a computational complexity of $\mathcal{O}(n)$. If we absorb history into the parameters to get $f_{W^*}$ and deduct with a limited context length, the deduction with discarded history has a reduced complexity of $\mathcal{O}(1)$, which suits scenarios with long sequence lengths.}
    \label{fig: computational cost}
\end{figure*}

\subsection{Memory issue of Transformers}
Transformers exhibit quadratic computational complexity $\mathcal{O}(N^2)$ due to their attention mechanism~\cite{vaswani2017attention}. Specifically, given a pretrained LLM $f$ with parameter $W$, generating a single token $x_{n+1} = f_W( x_1, \dots, x_{n})$ from an input sequence $X = (x_0, \dots, x_{n})$ costs $\mathcal{O}(n)$ with KV cache. Autoregressively, producing $N$ tokens results in a cumulative $\mathcal{O}(N^2)$ computational overhead and $\mathcal{O}(N)$ memory footprint (Fig.~\ref{fig: computational cost}). 
Numerous variants~\cite{child2019generating, beltagy2020longformer, katharopoulos2020transformers, choromanski2020rethinking, liu2023ring} have been proposed to resolve this memory issue, but they merely delay the inevitable exhaustion of cache memory and computing resources. 

\subsection{Superiority of parameter memory} The primary cause of the high computational cost of transformers in long contexts is their attention mechanism, which must process all historical contexts during deduction. To address this memory issue, prior works have proposed linear-time sequence models~\cite{peng2023rwkv, gu2021combining, gu2024mamba} that reduce the computational cost to $\mathcal{O}(N)$. 
Note that the linear-time models gain efficiency by sacrificing model capacity~\cite{merrill2024illusion}, as they compress arbitrarily rich history into a fixed-size hidden state $h_t \in \mathbb{R}^d$. By contrast, the paradigm that restores contexts into the model parameters is more promising, since the parameter space $\theta \in \mathbb{R}^{d \times d}$ is orders of magnitude larger than the activation space $h \in \mathbb{R}^d$, offering a vastly superior capacity for encoding history. Consequently, computationally costly contexts can be safely omitted from long-stream inferences. 

Formally, let $X$ denote the historical contexts and $Y$ denote the subsequent inferences. In the parameter memory paradigm, rather than performing standard inference over the full context $f_W(XY)$, we compute $f_{W^*}(Y)$ without the explicit context $X$, having stored the information of $X$ directly into the updated parameters $W^*$.
Such a parameter memory paradigm has been proposed by~\cite{sun2024learning, behrouz2025nested}.

\subsection{Analysis of existing parameter memory methods}
Test-Time Training (TTT)~\cite{sun2024learning, inplacettt}, the state-of-the-art parameter memory method, treats historical contexts as a continuous learning loop, updating model weights to memorize the incoming stream. That is:
\begin{equation}\label{eq: TTT optimization target}
    W^* = \arg\min_{W^*} \parallel f_{W^*}(x_\text{k})- x_\text{v} \parallel
\end{equation}
where $x_\text{k}$ and $x_\text{v}$ are conversions of input $x$, which build target projections. 
By training the model on target projections, TTT effectively converts context into training data and leverages the plasticity of neural networks to extend memory capacity, 
and reveals optimum performance in comprehensive context streams as batched training data. 
Overall, TTT is more suitable for scenarios of effective training, as it relies on adequate inputs to stabilize its training and has limitations when there are insufficient contexts that cannot form batch data, which is more common in the prevalent scenario of model inference in user interactions. 
Specifically, in the model inference process, TTT memorizes contexts by training the model to project them, and simply memorizing such inadequate data tokens often leads to the retention of irrelevant noise while failing to capture the high-level semantic dependencies required for reasoning~\cite{cao2025analyzing}. 
Furthermore, when projecting contexts, TTT does not specify the model's performance on subsequent inferences. This cannot invoke the causal mechanisms in LLM, as it does not maintain the contribution of removed historical sequences to future inferences, which is crucial for model deduction~\cite{geiger2021causal, geiger2024finding}. 
Overall, incorporating historical contexts is not just data summarization, but learning the causal representation~\cite{scholkopf2021toward} that maintains the intervention of history tokens to future ones.

\noindent \textbf{Motivation summary} To summarize,  we aim to reduce the $\mathcal{O}(N^2)$ computational cost of full-attention transformers by absorbing contexts into model parameters. Based on analyses of prior works, we propose to preserve the causal effects of historical context on subsequent sequences when building parameter memory.

\section{Proposed Method}

\subsection{Context absorption with causal relation synchronization}
Inspired by the parameter memory paradigm and the necessity of invoking causal mechanisms, we propose to move beyond context memorization to the preservation of the causal relation between historical contexts and the subsequent generation process. We define the required causal relation preservation in context memorization as the following self-supervised learning target: 
\textit{The updated contextless model $f_{W^*}$, with historical contexts $X$ absorbed into its parameters and access only to the subsequent texts $Y$, should behave identically to the original pretrained model $f_W$ with full contexts $XY$}. That is:
\begin{equation}\label{eq: absorber synchronization target}
    f_{W^*}(Y) \equiv f_W(XY)
\end{equation}
where $Y$ denotes the subsequent text following $X$, \textit{i.e.}, successive input tokens or model-generated inferences, such that the distribution of $Y$ is conditionally dependent on $X$. 
Note that such equivalence should be kept throughout the inference process of $Y$:
\begin{equation}\label{eq: absorber synchronization process}
    f_{W^*}(Y[:i]) = f_W(XY[:i]) \quad i=1, \dots, \vert Y \vert
\end{equation}
Then we get an optimization target:
\begin{equation}\label{eq: absorber optimization target}
    W^* = \arg\min_{W^*} 
    \sum_{i=1}^{\vert Y \vert}\parallel f_{W^*}(Y[:i]) - f_W(XY[:i]) \parallel
\end{equation}
Note that compared to prior works, our optimization is guided by the causal effect synchronization objective rather than simple reconstruction in Equation~\ref{eq: TTT optimization target}. Instead of trying to reconstruct the exact history $X$ token-by-token, our goal is to find a parameter configuration $W_*$ that preserves the causal effect of $X$ on the subsequent generation process of $Y$. This ensures that the model \textit{absorbs} existing contextual information and preserves its effects on further deduction, rather than simply memorizing it. Moreover, by applying semantic dependencies between $X$ and $Y$, such absorption only retains the relevant information in historical contexts $X$ for subsequent inference in $Y$, which filters noise in contextual data. By absorbing only the relevant information, our method relies less on high-quality contexts, and is more suitable for usual scenarios of model inferences.



\begin{algorithm}[ht]
\caption{Context Absorption}
\label{alg: context absorption}
\textbf{Inputs}: 
LLM $f_{W}$; Historical context $X=(x_1, x_2, \dots, x_n)$ to be absorbed into parameters $W$; Subsequent text segment $Y = (x_{n+1}, x_{n+1}, \dots, x_{n+m})$; Learning rate $\eta$; Maximum fine-tuning step number $K$; Synchronization threshold $\epsilon$ \\
\textbf{Outputs}:
Updated model parameters $W^*$ with context $X$ being absorbed.
\begin{algorithmic}[1]
    \STATE Forward $f_{W}(X Y)$, get full-context hidden states $H_1, \dots, H_{n}, H_{{n}+1}, \dots, H_{n+m}$
    \STATE Initialize $W^* \gets W$
    \FOR{$k \gets 1$ \textbf{to} $K$} 
        
        \STATE Forward $f_{W^*}(Y)$ to get contextless hidden states: $\tilde{H}_{{n}+1}, \dots, \tilde{H}_{n+m}$
        \STATE Compute synchronization loss: \\ $\mathcal{L} \gets \frac{1}{m} \sum_{p=n+1}^{n+m} \parallel \tilde{H}_p - {H}_p \parallel$ 
        \IF{$\mathcal{L}< \epsilon$ }
            \STATE \textbf{Return}{$W^*$} 
        \ENDIF
        \STATE $\nabla_{W^*}\mathcal{L} \gets \text{Backward}(\mathcal{L})$ 
        \STATE $W^* \gets W^* - \eta \cdot \nabla_{W^*}\mathcal{L}$ 
    \ENDFOR
    \STATE \textbf{Return}{$W^*$} 
\end{algorithmic}
\end{algorithm}
\vspace{-10pt}

\subsection{Absorb contexts through hidden states synchronization}
The goal of our method is to realize equations~\ref{eq: absorber synchronization target} through optimization in equation~\ref{eq: absorber optimization target} for causal effect synchronization. 
To ensure a more thorough synchronization, we force the hidden states of $f_{W^{*}}(Y)$ to be identical to $f_{W^*}(XY)$, which guarantees that they conduct identical deductions in the same internal manner. Specifically, let historical contexts $X=(x_1, x_2, \dots, x_n)$ be absorbed into model parameters, and we refer to future inferences $Y = (x_{n+1}, x_{n+1}, \dots, x_{n+m})$ to synchronize the behavior of the contextless model to the original full-context model; let $H_1, \dots, H_{n}, H_{{n}+1}, \dots, H_{n+m} $ be the hidden states in propagating $f_{W}(XY)$, $\tilde{H}_{{n}+1}, \dots, \tilde{H}_{m}$ be the hidden states in propagating $f_{W^*}(Y)$, where $H_{p} = (h_{p}^0, \dots, h_{p}^L), p=1, \dots, n+m$, and
$\tilde{H}_p = (\tilde{h}_p^0, \tilde{h}_p^1, \dots, \tilde{h}_p^L), p=n+1, \dots, n+m$, and $L$ is the number of layers in LLM $f$. Then the hidden states synchronization is:
\begin{equation}\label{eq: hiddenstate synchronization target}
    \tilde{H}_p = {H}_p \quad p=n+1, \dots, n+m
\end{equation}
which ensures the output synchronization in equation~\ref{eq: absorber synchronization process} by synchronizing the propagation process more thoroughly. Then we get the optimization target to ensure equation~\ref{eq: absorber optimization target}:
\begin{equation}\label{eq: hiddenstate synchronization optimization target}
    W^* = \arg \min_W \frac{1}{m} \sum_{p=n+1}^{n+m} \parallel \tilde{H}_p - {H}_p \parallel
\end{equation}
The above process that absorbs contexts $X$ into parameters $W$ is summarized in Algorithm~\ref{alg: context absorption}. Note that the absorption requires fine-tuning, and we employ LoRA~\cite{lora} for efficiency.

\subsection{Conduct absorption in the flowing sequence}
In a continual workflow where new texts, \textit{i.e.}, model generations or subsequent user prompts, are continuously absorbed into the model parameters, the synchronization process becomes iterative: 
Where new incoming contexts $X$ keep being absorbed into parameters $W$ by synchronization on subsequent texts $Y$. Specifically, we maintain a dynamic absorption window $Z = XY$. After the absorption of $X$, we move the window to absorb the segment subsequent to $X$. Such a continual absorption is detailed in Algorithm~\ref{alg:absorber_deduction}.

To summarize, we synchronize $f_{W^*}(Y)$ with $f_{W}(XY)$ using a fine-tuning strategy. By minimizing the divergence between the updated contextless model behavior and the ideal full-context behavior within the local segment, we treat the parameter update as a continuous error-correction process, which ensures the context absorption and generalization. Please refer to Figure~\ref{fig:overview} for a brief demonstration. Note that the computational cost of context absorption in Algorithm~\ref{alg: context absorption} is fixed given $m$ and $n$; in Algorithm~\ref{alg:absorber_deduction}, we absorb and generate $n$ tokens in each round, so that the computational costs grow linearly with sequence length. Overall, the computational cost of our method is $\mathcal{O}(N)$.

\begin{algorithm}[ht]
\caption{Absorber Deduction}
\label{alg:absorber_deduction}
\textbf{Inputs}: \\
Original pretrained LLM $f_{W}$; Input texts $I$; absorbed token number at one time: $n$; synchronization token number $m$ \\
\textbf{Outputs}: \\
Inference texts $S$
\begin{algorithmic}[1] 
    \STATE Initialize $t \gets 0$, absorption window $Z \gets I$, Inference texts $S = \varepsilon$
    \WHILE{\textbf{True}}
    \IF{$|Z| < m+n$}
        \FOR{$i \gets 1$ \textbf{to} $m+n-|Z|$}
            \STATE Forward $f_{W}(Z)$ to generate the next token $z$
            \IF{$z = \text{eos\_token}$}
                \STATE \textbf{Return} {$S$} 
            \ENDIF
            \STATE $Z \gets Zz, S \gets Sz$
        \ENDFOR
    \ENDIF
    \STATE $X \gets Z[:n], Y \gets Z[n+1:n+m]$
    \STATE Absorb $X$ into $W$ to get $W^*$ by synchronizing $f_{W^*}(Y)$ with $f_W(XY)$~(Algorithm~\ref{alg: context absorption})
    \STATE $Z \gets Z[n+1:]; W \gets W^*$
    \ENDWHILE
\end{algorithmic}
\end{algorithm}
\vspace{-10pt}

\section{Experiments}
\label{sec:experiments}

\subsection{Set up}

In this section, we evaluate the performance of {Absorber LLM} across four critical dimensions: computational efficiency, information retention in many-shot in-context learning, logical consistency in long-chain reasoning, and long text summarization.

\textbf{Settings.} Our experiments utilize the LLaMA2-7B~\cite{touvron2023llama} architecture as the primary backbone. All evaluations are conducted on an NVIDIA GeForce RTX 4080 SUPER GPU. In Algorithm~\ref{alg: context absorption}, we set $n=1024$ and $m=2048$. We employ the $L1$ norm for the loss function and use the AdamW~\cite{AdamW} optimizer with a learning rate of $\eta = 5\times 10^{-4}$. For LoRA, we set the rank to $r=64$, and the synchronization threshold is set to $\epsilon = 2$.

\textbf{Baselines.} We compare our method against standard LLaMA2-7B as a representative of full-attention transformers, Mamba \cite{gu2024mamba} as a representative SSM, and TTT \cite{sun2024learning} as a  parameter-memory baseline. 

\textbf{Datasets and Evaluation.}
For each of the four aforementioned tasks, we selected current popular datasets for evaluation. To assess computational efficiency, we employed a popular subset of the Pile called Books3~\cite{pile} and used the text generation time of the model as the evaluation metric. For information retention in many-shot in-context learning, we utilized the Agnews~\cite{Zhang2015CharacterlevelCN} dataset, which requires the model to classify different types of news articles. In our experiments, we provided the model with several news classification examples as the many-shot context, forming a long text input, and then asked the model to classify a new article. The classification accuracy served as the evaluation score for this task. 
To evaluate logical consistency in long-chain reasoning, we adopted the Musique~\cite{bai2023longbench} dataset, a collection of long-text reasoning problems. The model's task is to identify several relevant points from a lengthy passage and perform reasoning to derive the answer. 
The reasoning accuracy was used as the performance metric. Finally, for long text summarization, we employed the well-known SamSum~\cite{gliwa-etal-2019-samsum} dataset, which similarly tests the model's ability to comprehend and summarize extended texts. We computed the similarity between the model-generated summary and the reference summary using the Bleurt~\cite{sellam2020bleurt} model, and this similarity score served as the final evaluation metric.

\subsection{Inference Complexity and Latency Scaling}
We use Books3, a popular subset of the Pile. We assess the inference efficiency by measuring the average per-token latency ($\mathcal{L}$), which we define as the amortized time to generate $K=128$ tokens following a prefix of length $N$:
\begin{equation*}
    \mathcal{L}(N) = \frac{1}{K} \left( \mathcal{T}_{gen}(N+K) - \mathcal{T}_{prefill}(N) \right)
\end{equation*}
where $\mathcal{T}$ denotes wall-clock time. We vary the historical prefix $N$ from $10^3$ to $1.6 \times 10^4$ tokens. 

\begin{table}[H]
\caption{Comparison of amortized per-token latency $\mathcal{L}(N)$ across varying historical prefix lengths $N$. $K=128$ tokens are generated for each measurement.}
\label{tab:latency_comparison}
\centering
\resizebox{0.48\textwidth}{!}{

\begin{tabular}{lcccc}
\toprule
Method($10^{-3}$s) & $N=1\text{k} \sim 2\text{k}$ & $ N=2\text{k} \sim 4\text{k}$ & $N=4\text{k} \sim 8\text{k}$ &  $N=8 \text{k} \sim 16\text{k}$ \\
\midrule
Standard & 19.23 & 39.07 & 76.32 & OOM \\
Mamba &  \textbf{4.14} &  \textbf{4.52} &  \textbf{3.88} &  \textbf{3.69} \\
TTT & 20.33 & 20.39 & 19.16 & 19.31 \\
\midrule
Absorber & 29.73 & 30.22 & 30.38 & 29.98\\
\bottomrule
\end{tabular}
}
\end{table}
\vspace{-10pt}

As evidenced by the empirical data in Table~\ref{tab:latency_comparison}, the Standard Transformer exhibits a sharp, non-linear increase in inference latency as the context length scales, jumping from 19.23ms to 76.32ms, and eventually leading to an Out-of-Memory error beyond 8K tokens. This bottleneck stems from the quadratic complexity of self-attention and the mounting overhead of KV cache management. In contrast, Absorber LLM maintains a remarkably stable latency profile. While it carries a slightly higher initial overhead at short contexts, its latency remains near-constant regardless of the internalized context length. This demonstrates a practical $O(1)$ inference complexity.

\subsection{Long-Context In-Context Learning}
We evaluate many-shot In-Context Learning (ICL) performance using the Agnews dataset, focusing on tasks that require integrating multiple demonstration samples across long sequences. Table~\ref{tab:Agnews} summarizes the averaged scores across different sequence lengths.

\begin{table}[H]
\centering
\caption{Performance comparison on Agnews. We report macro-averaged Accuracy.}
\label{tab:Agnews}
\resizebox{0.45 \textwidth}{!}{
\begin{tabular}{lccc}
\toprule
Text Length  & 1k$~\sim$2k& 2k$~\sim$4k & 4k$~\sim$8k\\
\midrule
Standard &  \textbf{54.2} &  \textbf{49.8} & \textbf{43.7} \\
Mamba & 28.7 & 28.5 & 26.2 \\
TTT & 30.6 & 30.2 & 29.7 \\
\midrule
Absorber LLM & 37.9 & 38.2 & 35.5 \\
\bottomrule
\end{tabular}
}
\end{table}
\vspace{-10pt}

As shown in Table~\ref{tab:Agnews}, Absorber LLM consistently outperforms other linear-complexity baselines, namely Mamba and TTT, across all evaluated context windows. While the standard transformer maintains higher accuracy in shorter contexts ($<$4K), it imposes a heavy memory footprint within the 4K$\sim$8K range, limiting its practical deployment. Absorber LLM demonstrates a robust and stable performance profile, maintaining an accuracy of approximately $35\% \sim 38\%$, which is higher than the $28\% \sim 31\%$ range achieved by Mamba and TTT. These results indicate that our proposed functional alignment is more effective at distilling salient task-specific logic into model parameters than traditional reconstruction-based objectives.

\subsection{Logical Integrity in Multi-Step Reasoning}

In this part, we focus on multi-step deduction where the model must connect multiple factual links scattered across a long context. Specifically, we provide the model with the question and the initial reasoning steps, and task it with inferring the final answer based on this context. The accuracy of the answer is the score of the testing model.

\begin{table}[H]
\caption{Performance comparison on Musique. We report macro-averaged Accuracy.}
\label{tab:Musique}
\resizebox{0.48\textwidth}{!}{
\begin{tabular}{lccc}
\toprule
Method & 8k$~\sim$12k & 12k$~\sim$16k & 16k$~\sim$20k \\
\midrule
Standard & OOM & OOM & OOM \\
Standard-Truncation & 19.2 &  18.5 &  17.9 \\
Mamba & 31.5 & 30.4 & 27.6 \\
TTT & \textbf{34.3} & 30.9 & 28.4 \\
\midrule
Absorber LLM & 33.8 & \textbf{31.6} & \textbf{29.5} \\
\bottomrule
\end{tabular}
}
\end{table}
\vspace{-10pt}

As illustrated in Table \ref{tab:Musique}, the standard transformer fails to process ultra-long sequences ($>8K$) due to its quadratic memory complexity, resulting in OOM errors. To provide a viable baseline, we introduce standard-truncation, a method that constrains the standard model by truncating the input to its maximum supported context window, effectively discarding earlier historical information to avoid memory exhaustion. The experimental results demonstrate that as the sequence length scales to the ultra-long regime ($16K \sim 20K$), the performance of other linear-complexity models like Mamba and TTT degrades significantly, dropping to 27.6 and 28.4, respectively. Notably, Absorber LLM exhibits superior scalability, maintaining a higher accuracy of 29.5.

\subsection{Long Text Summary}

In this experiment, we selected three different lengths of text, namely less than 2k$\sim$4k, 4k$\sim$8k, and 8k$\sim$16k. For each length of text, we used different methods to input the text and extract the model output. Finally, we compared the model output with the standard answers in the Samsum dataset, and used the Bleurt model~\cite{sellam2020bleurt} to calculate the similarity between the two as the final score. 
\\

\begin{table}[H]
\caption{Performance of the model on Samsum of different text lengths}
\label{tab:longbench3}
\centering
\resizebox{0.45\textwidth}{!}{
\begin{tabular}{lccc}
\toprule
Text Length & 2k$\sim$4k & 4k$\sim$8k  & 8k$\sim$16k \\
\midrule
Standard &  \textbf{0.433} &  \textbf{0.428} & OOM \\
Mamba & 0.428 & 0.413 & 0.395 \\
TTT & 0.415 & 0.406 & 0.391 \\
\midrule
Absorber LLM & 0.423& 0.417 & \textbf{0.408} \\
\bottomrule
\end{tabular}
}

\end{table}
\vspace{-10pt}

In the summarization task, we evaluate the model's ability to capture salient information from long dialogues. As shown in Table \ref{tab:longbench3}, the standard transformer serves as the performance upper bound in shorter contexts; however, its memory requirements become prohibitive as the sequence scales, eventually leading to an OOM error in the $8K \sim 16K$ range.In the $2K \sim 4K$ range, while Absorber LLM (0.423) still trails the standard model (0.433), it significantly outperforms the other linear-complexity baselines, exceeding Mamba and TTT by 0.008 and 0.005 points, respectively. This suggests that in the early stages of context extension, functional alignment provides a more effective compression mechanism for dialogue structures than the hidden state updates in Mamba or the reconstruction objective in TTT. As the text length increases to $8K \sim 16K$, the advantage of Absorber LLM as a scalable alternative becomes decisive. While the Standard model is no longer functional due to memory exhaustion, Absorber LLM maintains a stable score of 0.408, outperforming Mamba (0.395) by a clear margin of 0.013. This indicates that Absorber LLM successfully internalizes global dialogue dependencies into its weights. 

\subsection{Ablation Study}

\subsubsection{F1 Score}
The F1 score is defined as the harmonic mean of precision and recall:
\begin{equation}
F_1 = 2 \cdot \frac{\text{Precision} \cdot \text{Recall}}{\text{Precision} + \text{Recall}}
\end{equation}
In this context, \textit{Precision} is the ratio of the number of shared tokens to the total number of tokens in the predicted sequence, while \textit{Recall} is the ratio of the number of shared tokens to the total number of tokens in the ground truth sequence. By treating both the prediction and the reference as bags of tokens, the F1 score effectively captures the semantic proximity and the quality of the generated reasoning steps.

\subsubsection{hidden states and token}
To verify the necessity of layer-wise functional alignment, we conduct an ablation study comparing our approach, aligning internal hidden states, against a baseline that aligns only the token-level output distributions, \textit{i.e.}, logits. In the token-level alignment variant, we replace the state-wise MSE loss with the KL divergence loss between the next-token probabilities of the contextless model and the full-context oracle.

\begin{table}[H]
\caption{Ablation results on alignment granularity using Samsum. Token Alignment constraints only the output layer, while Hidden State Alignment constraints the entire representational space.}
\label{tab:granularity}
\centering
\resizebox{0.4\textwidth}{!}{
\begin{tabular}{lcc}
\toprule
Alignment Target & F1 Score & BRT Score\\
\midrule
Token Distribution & 37.4 & 0.324 \\
Hidden States & \textbf{57.5} & \textbf{0.415} \\
\bottomrule
\end{tabular}
}

\end{table}
\vspace{-10pt}

The results in Table \ref{tab:granularity} reveal a significant performance gap. We observe that token-level alignment alone is insufficient for Samsum. We believe that this shallow supervision only optimizes the final predictive head, failing to internalize the complex causal structures. In contrast, by forcing the contextless model's hidden states to match the oracle, we ensure that the high-dimensional semantic space, which has a vastly superior capacity for encoding history compared to surface-level tokens, is fully utilized. This deep alignment effectively preserves the deductive mechanisms required for reasoning and summarizing, preventing the model from collapsing into simple statistical pattern matching.

\subsubsection{hyperparameter experiment on $n$ and $m$}

We investigate the impact of the absorption segment length $n$ and the behavioral alignment window $m$ on the model's performance. Table \ref{tab:ablation} presents the accuracy on a validation subset of Samsum. We observe that increasing the reference length $m$ consistently improves alignment quality by providing a broader supervision signal for the model's behavior. Our results indicate that performance generally improves as the absorption segment length $n$ increases, because a larger $n$ enables the model to capture more holistic contextual dependencies and internalize comprehensive semantic structures within each update. Consequently, the configuration with a larger $n$ combined with a larger $m$ provides an optimal balance for maintaining high alignment fidelity during long-stream inference.

\begin{table}[H]
\caption{Ablation study on internalization segment length $n$ and behavioral alignment window $m$. Scores represent Accuracy (\%).}
\label{tab:ablation}
\resizebox{0.48\textwidth}{!}{
\begin{tabular}{c|ccc}
\toprule
Accuracy (\%) & $m=512$ & $m=1024$ & $m=2048$ \\
\midrule
$n=256$  & 35.1 & 57.9 & 42.3 \\
$n=512$  & 41.9 & 46.5 & 55.1 \\
$n=1024$ & 47.8 & 50.8 & \textbf{61.9} \\
\bottomrule
\end{tabular}
}
\end{table}
\vspace{-10pt}

\subsubsection{Impact of Regularization}

We investigate the influence of different regularization strategies on the parameter absorption process. In our framework, the weight update involves a trade-off between minimizing the functional alignment loss and maintaining the structural integrity of the pretrained parameters. We compare two configurations: (1) alignment with L1-norm regularization to promote sparsity in updates, and (2) alignment with L2-norm regularization to encourage smooth parameter transitions. The experiments are conducted on Samsum dataset to assess how these constraints affect long-range logical deduction.

\begin{table}[H]
\caption{Ablation study on regularization methods for context absorption. Results are reported on the Samsum dataset.}
\label{tab:regularization}
\centering
\resizebox{0.4\textwidth}{!}{
\begin{tabular}{lcc}
\toprule
Regularization & F1 Score & BRT Score \\
\midrule
L1 Regularization & \textbf{57.5} & \textbf{0.415} \\
L2 Regularization & 42.3 & 0.383 \\
\bottomrule
\end{tabular}
}
\end{table}
\vspace{-10pt}

The empirical findings presented in Table~\ref{tab:regularization} indicate that L1 regularization is superior for the context absorption paradigm. We observe that L2 regularization, by imposing a uniform penalty on all weight magnitudes, tends to diffusely suppress updates, which may hinder the formation of the sharp, specific parameter changes required to encode discrete logical relations. This leads to a noticeable drop in both F1 and BRT scores. L1 regularization, by inducing sparsity, acts as a selective constraint. It effectively preserves the structural integrity of the pretrained representational space while providing sufficient plasticity to internalize sparse causal dependencies. This balance is crucial for maintaining the deductive capability of transformers.

\section{Related Work}

\paragraph{Efficient Transformers and Linear Attention.}
Standard self-attention computes the output as $O = \text{softmax}(QK^T/\sqrt{d})V$, incurring $\mathcal{O}(N^2)$ computational complexity due to the $N \times N$ attention matrix. Linear attention methods~\cite{katharopoulos2020transformers, choromanski2020rethinking} bypass this by employing a kernel trick. By approximating the softmax operation with feature maps $\phi(\cdot)$ and exploiting the associativity of matrix multiplication, the attention mechanism becomes:
\begin{equation*}
    O_i = \phi(Q_i)^T \sum_{j=1}^{i} \left( \phi(K_j) V_j^T \right)
\end{equation*}
Here, the term $\sum \phi(K_j) V_j^T$ acts as a matrix-valued hidden state $S_i \in \mathbb{R}^{d \times d}$, updated recurrently as $S_i = S_{i-1} + \phi(K_i)V_i^T$. This reduces inference complexity to $\mathcal{O}(N)$. However, strictly adhering to this linear recurrence degrades retrieval capabilities compared to full attention, primarily due to its deviation from the exact softmax mechanism.

\paragraph{State Space Models (SSMs) and Modern RNNs.}
Recent architectures like Mamba~\cite{gu2024mamba} and RWKV~\cite{peng2023rwkv} discretize continuous-time state space equations into a recurrent form. The core mechanism involves compressing history into a hidden state $h_t \in \mathbb{R}^d$ by:
\begin{equation*}
    h_t = \bar{A}_t h_{t-1} + \bar{B}_t x_t, \quad y_t = C h_t
\end{equation*}
where $\bar{A}_t$ and $\bar{B}_t$ are discretized parameters (which, in the case of Mamba, are data-dependent). While efficient ($\mathcal{O}(1)$ inference state), these models are constrained by the Fixed-Capacity Hypothesis~\cite{merrill2024illusion}. The state $h_t$ must encode the entire context $x_{1:t}$. Since $\text{dim}(h_t) \ll \text{dim}(x_{1:t})$ for long sequences, this inevitably results in lossy compression and the catastrophic forgetting of long-tail dependencies that cannot be fit into the bounded vector space $\mathbb{R}^d$.

\paragraph{Parametric Memory and Test-Time Training (TTT).}
To expand memory capacity beyond fixed vectors, TTT~\cite{sun2024learning} adopts model parameters as the storage medium. TTT formulates memory retrieval as an online self-supervised learning problem. Given a history sequence or a new token $x_t$), it updates the internal weights $W_{t-1}$ to $W_t$ by minimizing a reconstruction objective $\ell$ in Equation~\ref{eq: TTT optimization target}:
\begin{equation*}
    W_t = W_{t-1} - \eta \nabla_W \ell(W_{t-1}, x_t)
\end{equation*}
This process effectively embeds the context into the high-dimensional parameter space $\mathbb{R}^{d \times d}$, theoretically offering vastly superior memory capacity compared to vector states. However, this approach relies on iterative approximation through Stochastic Gradient Descent~(SGD). Consequently, the quality of memory retention becomes heavily dependent on proper optimization hyperparameters and adequate training data. 
Furthermore, when projecting contexts, TTT does not specify the model's performance on subsequent inferences. This cannot invoke the causal mechanisms in LLM, as it does not maintain the contribution of removed historical sequences to future inferences. 

\paragraph{Predictive Information and Causal Alignment.}
Our formulation of parameter memory moves beyond simple target projection toward functional alignment, proposing a concept that bridges information theory and causal inference.
From an information-theoretic perspective, the information bottleneck principle~\cite{tishby2000information, bialek2001predictability} suggests that an optimal representation should discard historical details irrelevant to predicting the future. 
More fundamentally, this aligns with recent advances in causal abstraction~\cite{geiger2021causal, geiger2024finding}, which posit that a simpler compressed model serves as a valid surrogate for a complex full system only if it preserves the causal mechanisms governing the output. In this view, context compression is not merely data summarization, but rather the learning of an invariant causal representation~\cite{scholkopf2021toward} that maintains the intervention-effect relationship between historical and future tokens.
While prior works in prompt compression~\cite{chevalier2023adapting, mu2023learning} apply this philosophy at the token level, Absorber LLM extends causal alignment to the parameter space. By enforcing our equivalence condition, we ensure that the compressed parameters functionally substitute the full history, preserving the causal dynamics of generation without retaining the raw data.

\section{Future Work}
In this paper, we presented Absorber LLM, a framework that internalizes historical context into model parameters by enforcing functional alignment. Our proposed framework can be extended along three main directions. 

First, we will explore how context absorption can be integrated with continual learning, enabling models to incrementally internalize new data while avoiding catastrophic forgetting of previously absorbed memories. Second, we will investigate the causal editability of absorbed representations to support selective forgetting, allowing specific memories to be removed for privacy or factual updates without full retraining. Third, we aim to extend the absorption paradigm to multimodal long-context settings, such as long-form video and sensor streams, where functional alignment could offer a scalable alternative to expanding the KV cache. 

These extensions will advance the development of more efficient, controllable, and broadly capable long-context models. We plan to optimize and expand the framework to make it applicable to longer texts and larger-parameter models.
\section{Conclusion}

This paper introduced Absorber LLM, which internalizes long contexts into model parameters through functional alignment, moving beyond the computational constraints of conventional transformer KV-caches. By aligning layer-wise representations between a contextless model and a full-context oracle, our approach encodes semantic dependencies directly into the parameter space, enabling causal and coherent reasoning without expanding external memory. Experiments demonstrate that Absorber LLM achieves constant-time inference and outperforms existing methods on long-context benchmarks, while robustly handling dynamic updates.

\bibliography{references}

\begin{thebibliography}{35}
\providecommand{\natexlab}[1]{#1}
\providecommand{\url}[1]{\texttt{#1}}
\expandafter\ifx\csname urlstyle\endcsname\relax
  \providecommand{\doi}[1]{doi: #1}\else
  \providecommand{\doi}{doi: \begingroup \urlstyle{rm}\Url}\fi

\bibitem[Bai et~al.(2023)Bai, Lv, Zhang, Lyu, Tang, Huang, Du, Liu, Zeng, Hou, Dong, Tang, and Li]{bai2023longbench}
Bai, Y., Lv, X., Zhang, J., Lyu, H., Tang, J., Huang, Z., Du, Z., Liu, X., Zeng, A., Hou, L., Dong, Y., Tang, J., and Li, J.
\newblock Longbench: A bilingual, multitask benchmark for long context understanding, 2023.

\bibitem[Bairi et~al.(2024)Bairi, Sonwane, Kanade, C, Iyer, Parthasarathy, Rajamani, Ashok, and Shet]{bairi2024codeplan}
Bairi, R., Sonwane, A., Kanade, A., C, V.~D., Iyer, A., Parthasarathy, S., Rajamani, S., Ashok, B., and Shet, S.
\newblock Codeplan: Repository-level coding using llms and planning.
\newblock \emph{Proceedings of the ACM on Software Engineering}, 1\penalty0 (FSE):\penalty0 675--698, 2024.

\bibitem[Behrouz et~al.()Behrouz, Razaviyayn, Zhong, and Mirrokni]{behrouz2025nested}
Behrouz, A., Razaviyayn, M., Zhong, P., and Mirrokni, V.
\newblock Nested learning: The illusion of deep learning architectures.
\newblock In \emph{The Thirty-ninth Annual Conference on Neural Information Processing Systems}.

\bibitem[Beltagy et~al.(2020)Beltagy, Peters, and Cohan]{beltagy2020longformer}
Beltagy, I., Peters, M.~E., and Cohan, A.
\newblock Longformer: The long-document transformer.
\newblock \emph{arXiv preprint arXiv:2004.05150}, 2020.

\bibitem[Bialek et~al.(2001)Bialek, Nemenman, and Tishby]{bialek2001predictability}
Bialek, W., Nemenman, I., and Tishby, N.
\newblock Predictability, complexity, and learning.
\newblock \emph{Neural computation}, 13\penalty0 (11):\penalty0 2409--2463, 2001.

\bibitem[Cao et~al.(2025)Cao, Schooler, and Zafarani]{cao2025analyzing}
Cao, Z., Schooler, L., and Zafarani, R.
\newblock Analyzing memory effects in large language models through the lens of cognitive psychology.
\newblock \emph{arXiv preprint arXiv:2509.17138}, 2025.

\bibitem[Chevalier et~al.(2023)Chevalier, Wettig, Ajith, and Chen]{chevalier2023adapting}
Chevalier, A., Wettig, A., Ajith, A., and Chen, D.
\newblock Adapting language models to compress contexts.
\newblock In \emph{Proceedings of the 2023 Conference on Empirical Methods in Natural Language Processing}, pp.\  3829--3846, 2023.

\bibitem[Child(2019)]{child2019generating}
Child, R.
\newblock Generating long sequences with sparse transformers.
\newblock \emph{arXiv preprint arXiv:1904.10509}, 2019.

\bibitem[Choromanski et~al.(2021)Choromanski, Likhosherstov, Dohan, Song, Gane, Sarlos, Hawkins, Davis, Mohiuddin, Kaiser, Belanger, Colwell, and Weller]{choromanski2020rethinking}
Choromanski, K.~M., Likhosherstov, V., Dohan, D., Song, X., Gane, A., Sarlos, T., Hawkins, P., Davis, J.~Q., Mohiuddin, A., Kaiser, L., Belanger, D.~B., Colwell, L.~J., and Weller, A.
\newblock Rethinking attention with performers.
\newblock In \emph{International Conference on Learning Representations}, 2021.
\newblock URL \url{https://openreview.net/forum?id=Ua6zuk0WRH}.

\bibitem[Dao et~al.(2022)Dao, Fu, Ermon, Rudra, and R{\'e}]{dao2022flashattention}
Dao, T., Fu, D., Ermon, S., Rudra, A., and R{\'e}, C.
\newblock Flashattention: Fast and memory-efficient exact attention with io-awareness.
\newblock \emph{Advances in neural information processing systems}, 35:\penalty0 16344--16359, 2022.

\bibitem[Feng et~al.(2026)Feng, Luo, Hua, Zhang, Huang, He, and Cai]{inplacettt}
Feng, G., Luo, S., Hua, K., Zhang, G., Huang, W., He, D., and Cai, T.
\newblock In-place test-time training.
\newblock In \emph{The Fourteenth International Conference on Learning Representations}, 2026.
\newblock URL \url{https://openreview.net/forum?id=dTWfCLSoyl}.

\bibitem[Gao et~al.(2020)Gao, Biderman, Black, Golding, Hoppe, Foster, Phang, He, Thite, Nabeshima, Presser, and Leahy]{pile}
Gao, L., Biderman, S., Black, S., Golding, L., Hoppe, T., Foster, C., Phang, J., He, H., Thite, A., Nabeshima, N., Presser, S., and Leahy, C.
\newblock The {P}ile: An 800gb dataset of diverse text for language modeling.
\newblock \emph{arXiv preprint arXiv:2101.00027}, 2020.

\bibitem[Geiger et~al.(2021)Geiger, Lu, Icard, and Potts]{geiger2021causal}
Geiger, A., Lu, H., Icard, T., and Potts, C.
\newblock Causal abstractions of neural networks.
\newblock \emph{Advances in Neural Information Processing Systems}, 34:\penalty0 9574--9586, 2021.

\bibitem[Geiger et~al.(2024)Geiger, Wu, Potts, Icard, and Goodman]{geiger2024finding}
Geiger, A., Wu, Z., Potts, C., Icard, T., and Goodman, N.
\newblock Finding alignments between interpretable causal variables and distributed neural representations.
\newblock In \emph{Causal Learning and Reasoning}, pp.\  160--187. PMLR, 2024.

\bibitem[Gliwa et~al.(2019)Gliwa, Mochol, Biesek, and Wawer]{gliwa-etal-2019-samsum}
Gliwa, B., Mochol, I., Biesek, M., and Wawer, A.
\newblock {SAMS}um corpus: A human-annotated dialogue dataset for abstractive summarization.
\newblock In \emph{Proceedings of the 2nd Workshop on New Frontiers in Summarization}, pp.\  70--79, Hong Kong, China, November 2019. Association for Computational Linguistics.
\newblock \doi{10.18653/v1/D19-5409}.
\newblock URL \url{https://www.aclweb.org/anthology/D19-5409}.

\bibitem[Gu \& Dao(2024)Gu and Dao]{gu2024mamba}
Gu, A. and Dao, T.
\newblock Mamba: Linear-time sequence modeling with selective state spaces.
\newblock In \emph{First conference on language modeling}, 2024.

\bibitem[Gu et~al.(2021)Gu, Johnson, Goel, Saab, Dao, Rudra, and R{\'e}]{gu2021combining}
Gu, A., Johnson, I., Goel, K., Saab, K., Dao, T., Rudra, A., and R{\'e}, C.
\newblock Combining recurrent, convolutional, and continuous-time models with linear state space layers.
\newblock \emph{Advances in neural information processing systems}, 34:\penalty0 572--585, 2021.

\bibitem[Hsieh et~al.(2024)Hsieh, Sun, Kriman, Acharya, Rekesh, Jia, and Ginsburg]{hsieh2024ruler}
Hsieh, C.-P., Sun, S., Kriman, S., Acharya, S., Rekesh, D., Jia, F., and Ginsburg, B.
\newblock {RULER}: What{\textquoteright}s the real context size of your long-context language models?
\newblock In \emph{First Conference on Language Modeling}, 2024.
\newblock URL \url{https://openreview.net/forum?id=kIoBbc76Sy}.

\bibitem[Hu et~al.(2022)Hu, Shen, Wallis, Allen-Zhu, Li, Wang, Wang, Chen, et~al.]{lora}
Hu, E.~J., Shen, Y., Wallis, P., Allen-Zhu, Z., Li, Y., Wang, S., Wang, L., Chen, W., et~al.
\newblock Lora: Low-rank adaptation of large language models.
\newblock \emph{ICLR}, 1\penalty0 (2):\penalty0 3, 2022.

\bibitem[Jimenez et~al.(2024)Jimenez, Yang, Wettig, Yao, Pei, Press, and Narasimhan]{jimenez2023swe}
Jimenez, C.~E., Yang, J., Wettig, A., Yao, S., Pei, K., Press, O., and Narasimhan, K.~R.
\newblock {SWE}-bench: Can language models resolve real-world github issues?
\newblock In \emph{The Twelfth International Conference on Learning Representations}, 2024.
\newblock URL \url{https://openreview.net/forum?id=VTF8yNQM66}.

\bibitem[Katharopoulos et~al.(2020)Katharopoulos, Vyas, Pappas, and Fleuret]{katharopoulos2020transformers}
Katharopoulos, A., Vyas, A., Pappas, N., and Fleuret, F.
\newblock Transformers are rnns: Fast autoregressive transformers with linear attention.
\newblock In \emph{International conference on machine learning}, pp.\  5156--5165. PMLR, 2020.

\bibitem[Liu et~al.(2024)Liu, Zaharia, and Abbeel]{liu2023ring}
Liu, H., Zaharia, M., and Abbeel, P.
\newblock Ringattention with blockwise transformers for near-infinite context.
\newblock In \emph{The Twelfth International Conference on Learning Representations}, 2024.
\newblock URL \url{https://openreview.net/forum?id=WsRHpHH4s0}.

\bibitem[Loshchilov \& Hutter(2019)Loshchilov and Hutter]{AdamW}
Loshchilov, I. and Hutter, F.
\newblock Decoupled weight decay regularization.
\newblock In \emph{International Conference on Learning Representations}, 2019.
\newblock URL \url{https://openreview.net/forum?id=Bkg6RiCqY7}.

\bibitem[Merrill et~al.(2024)Merrill, Petty, and Sabharwal]{merrill2024illusion}
Merrill, W., Petty, J., and Sabharwal, A.
\newblock The illusion of state in state-space models.
\newblock In \emph{Forty-first International Conference on Machine Learning}, 2024.
\newblock URL \url{https://openreview.net/forum?id=QZgo9JZpLq}.

\bibitem[Mu et~al.(2023)Mu, Li, and Goodman]{mu2023learning}
Mu, J., Li, X., and Goodman, N.
\newblock Learning to compress prompts with gist tokens.
\newblock \emph{Advances in Neural Information Processing Systems}, 36:\penalty0 19327--19352, 2023.

\bibitem[Park et~al.(2023)Park, O'Brien, Cai, Morris, Liang, and Bernstein]{park2023generative}
Park, J.~S., O'Brien, J., Cai, C.~J., Morris, M.~R., Liang, P., and Bernstein, M.~S.
\newblock Generative agents: Interactive simulacra of human behavior.
\newblock In \emph{Proceedings of the 36th Annual ACM Symposium on User Interface Software and Technology}, pp.\  1--22, 2023.

\bibitem[Peng et~al.(2023)Peng, Alcaide, Anthony, Albalak, Arcadinho, Biderman, Cao, Cheng, Chung, Grella, et~al.]{peng2023rwkv}
Peng, B., Alcaide, E., Anthony, Q., Albalak, A., Arcadinho, S., Biderman, S., Cao, H., Cheng, X., Chung, M., Grella, M., et~al.
\newblock Rwkv: Reinventing rnns for the transformer era.
\newblock \emph{arXiv preprint arXiv:2305.13048}, 2023.

\bibitem[Sch{\"o}lkopf et~al.(2021)Sch{\"o}lkopf, Locatello, Bauer, Ke, Kalchbrenner, Goyal, and Bengio]{scholkopf2021toward}
Sch{\"o}lkopf, B., Locatello, F., Bauer, S., Ke, N.~R., Kalchbrenner, N., Goyal, A., and Bengio, Y.
\newblock Toward causal representation learning.
\newblock \emph{Proceedings of the IEEE}, 109\penalty0 (5):\penalty0 612--634, 2021.

\bibitem[Sellam et~al.(2020)Sellam, Das, and Parikh]{sellam2020bleurt}
Sellam, T., Das, D., and Parikh, A.
\newblock Bleurt: Learning robust metrics for text generation.
\newblock In \emph{Proceedings of the 58th Annual Meeting of the Association for Computational Linguistics}. Association for Computational Linguistics, 2020.

\bibitem[Sun et~al.(2025)Sun, Li, Dalal, Xu, Vikram, Zhang, Dubois, Chen, Wang, Koyejo, Hashimoto, and Guestrin]{sun2024learning}
Sun, Y., Li, X., Dalal, K., Xu, J., Vikram, A., Zhang, G., Dubois, Y., Chen, X., Wang, X., Koyejo, S., Hashimoto, T., and Guestrin, C.
\newblock Learning to (learn at test time): {RNN}s with expressive hidden states.
\newblock In \emph{Forty-second International Conference on Machine Learning}, 2025.
\newblock URL \url{https://openreview.net/forum?id=wXfuOj9C7L}.

\bibitem[Tishby et~al.(2000)Tishby, Pereira, and Bialek]{tishby2000information}
Tishby, N., Pereira, F.~C., and Bialek, W.
\newblock The information bottleneck method.
\newblock \emph{arXiv preprint physics/0004057}, 2000.

\bibitem[Touvron et~al.(2023)Touvron, Lavril, Izacard, Martinet, Lachaux, Lacroix, Rozi{\`e}re, Goyal, Hambro, Azhar, Rodriguez, Joulin, Grave, and Lample]{touvron2023llama}
Touvron, H., Lavril, T., Izacard, G., Martinet, X., Lachaux, M.-A., Lacroix, T., Rozi{\`e}re, B., Goyal, N., Hambro, E., Azhar, F., Rodriguez, A., Joulin, A., Grave, E., and Lample, G.
\newblock Llama: Open and efficient foundation language models.
\newblock \emph{arXiv preprint arXiv:2302.13971}, 2023.

\bibitem[Vaswani et~al.(2017)Vaswani, Shazeer, Parmar, Uszkoreit, Jones, Gomez, Kaiser, and Polosukhin]{vaswani2017attention}
Vaswani, A., Shazeer, N., Parmar, N., Uszkoreit, J., Jones, L., Gomez, A.~N., Kaiser, {\L}., and Polosukhin, I.
\newblock Attention is all you need.
\newblock \emph{Advances in neural information processing systems}, 30, 2017.

\bibitem[Wang et~al.(2024)Wang, Ma, Feng, Zhang, Yang, Zhang, Chen, Tang, Chen, Lin, et~al.]{wang2024survey}
Wang, L., Ma, C., Feng, X., Zhang, Z., Yang, H., Zhang, J., Chen, Z., Tang, J., Chen, X., Lin, Y., et~al.
\newblock A survey on large language model based autonomous agents.
\newblock \emph{Frontiers of Computer Science}, 18\penalty0 (6):\penalty0 186345, 2024.

\bibitem[Zhang et~al.(2015)Zhang, Zhao, and LeCun]{Zhang2015CharacterlevelCN}
Zhang, X., Zhao, J.~J., and LeCun, Y.
\newblock Character-level convolutional networks for text classification.
\newblock In \emph{NIPS}, 2015.

\end{thebibliography}
\bibliographystyle{icml2026/icml2026}




\end{document}